\documentclass[letterpaper, 10 pt, conference]{ieeeconf}
\IEEEoverridecommandlockouts    % This command is only needed if
\usepackage{graphics}           
\usepackage{times}              
\usepackage{amsmath}            
\usepackage{amssymb}            
\usepackage{graphicx}
\usepackage{algorithm}
\usepackage[noend]{algpseudocode}
\usepackage{booktabs}
\usepackage{color}
\definecolor{instructioncolor}{rgb}{.5,.5,.5}

\usepackage[font=small]{caption}

\def\secref#1{Sec.~\ref{#1}}
\def\figref#1{Fig.~\ref{#1}}
\def\tabref#1{Tab.~\ref{#1}}
\def\eqref#1{Eq.~(\ref{#1})}

\makeatletter
\usepackage{xspace}
\DeclareRobustCommand\onedot{\futurelet\@let@token\@onedot}
\def\@onedot{\ifx\@let@token.\else.\null\fi\xspace}

\def\etal{{et al}\onedot}
\makeatother

\usepackage{array}
\newcolumntype{L}[1]{>{\raggedright\let\newline\\\arraybackslash\hspace{0pt}}m{#1}}
\newcolumntype{C}[1]{>{\centering\let\newline\\\arraybackslash\hspace{0pt}}m{#1}}
\newcolumntype{R}[1]{>{\raggedleft\let\newline\\\arraybackslash\hspace{0pt}}m{#1}}
\usepackage{hyperref}
\usepackage{subfigure}
\usepackage[dvipsnames]{xcolor}
\title{\LARGE \bf SHINE-Mapping: Large-Scale 3D Mapping \\ Using Sparse Hierarchical Implicit Neural Representations}

\author{Xingguang Zhong$^*$ \and Yue Pan$^*$ \and Jens Behley \and Cyrill Stachniss% <-this % stops a space
  \thanks{$^*$ Equal contribution.}
  \thanks{All authors are with the University of Bonn, Germany. Cyrill Stachniss is additionally with the Department of Engineering Science at the University of Oxford, UK, and with the Lamarr Institute for Machine Learning and Artificial Intelligence, Germany.}%
  \thanks{This work has partially been funded by the European Union under the grant agreements No~101070405~(DigiForest) and No~101017008~(Harmony).
  }%
}

\begin{document}
\maketitle
\thispagestyle{empty}
\pagestyle{empty}

%%%%%%%%%%%%%%%%%%%%%%%%%%%%%%%%%%%%%%%%%%%%%%%%%%%%%%%%%%%%%%%%%%%%%%%%%%%%%%%%
\begin{abstract}
Accurate mapping of large-scale environments is an essential building block of most outdoor autonomous systems. Challenges of traditional mapping methods include the balance between memory consumption and mapping accuracy. This paper addresses the problem of achieving large-scale 3D reconstruction using implicit representations built from 3D LiDAR measurements. We learn and store implicit features through an octree-based, hierarchical structure, which is sparse and extensible. The implicit features can be turned into signed distance values through a shallow neural network. We leverage binary cross entropy loss to optimize the local features with the 3D measurements as supervision. Based on our implicit representation, we design an incremental mapping system with regularization to tackle the issue of forgetting in continual learning. Our experiments show that our 3D reconstructions are more accurate, complete, and memory-efficient than current state-of-the-art 3D mapping methods.
\end{abstract}

%%%%%%%%%%%%%%%%%%%%%%%%%%%%%%%%%%%%%%%%%%%%%%%%%%%%%%%%%%%%%%%%%%%%%%%%%%%%%%%%
\section{Introduction}
\label{sec:intro}

Localization and navigation in large-scale outdoor scenes is a common task of mobile robots, especially for self-driving cars. An accurate and dense 3D map of the environment plays a relevant role in achieving these tasks, and most systems use or maintain a model of their surroundings. Usually, for outdoors, dense 3D maps are built based on range sensors such as 3D LiDAR~\cite{behley2018rss, newcombe2011ismar}.
Due to the limited memory of most mobile robots performing all computations onboard, maps should be compact but, at the same time, accurate enough to represent the environment in sufficient detail. 

Current large-scale mapping methods often use spatial grids or various tree structures as map representation~\cite{hornung2013ar, kuehner2020icra, milan2018icra, pan2022iros, vizzo2022sensors}. For these models, it can be hard to simultaneously satisfy both desires, accurate and detailed 3D information but not requiring substantial memory resources. Additionally, these methods often do not perform well in areas that have been only sparsely covered with sensor data. In such areas, they usually cannot reconstruct a map at a high level of completeness.

Recently, neural network-based representations~\cite{mescheder2019cvpr, mildenhall2020eccv, park2019cvpr} attracted significant interest in the computer vision and robotics communities. By storing information about the environment in the neural network implicitly, these approaches can achieve remarkable accuracy and high-fidelity reconstructions using a comparably compact representation. Due to such advantages, several recent works have used implicit representation for 3D scene reconstruction built from images data or RGB-D frames~\cite{azinovic2022cvpr, sucar2021iccv, sun2021cvpr, wang2022threedv, zhu2022cvpr}. Comparably, little has been done in the context of LiDAR data. Furthermore, most of these methods only work in relatively small indoor scenes, which is difficult to use in robot applications for large-scale outdoor environments.

In this paper, we address this challenge and investigate effective means of using an implicit neural network-based map representation for outdoor robots using range sensors such as LiDARs.
%Based on 3D point clouds and robot poses, we are looking for an effective way to map and reconstruct large outdoor environments using implicit representations. 
%There are several ways how to approach this challenge. 
We took inspiration from the recent work by Takikawa \etal~\cite{takikawa2021cvpr}, which represents surfaces with a sparse octree-based feature volume and can adaptively fit 3D shapes with multiple levels of detail. 
% be more concise and to the point here in the introduction

\begin{figure}[tb]
  \centering
  \includegraphics[width=0.8\linewidth]{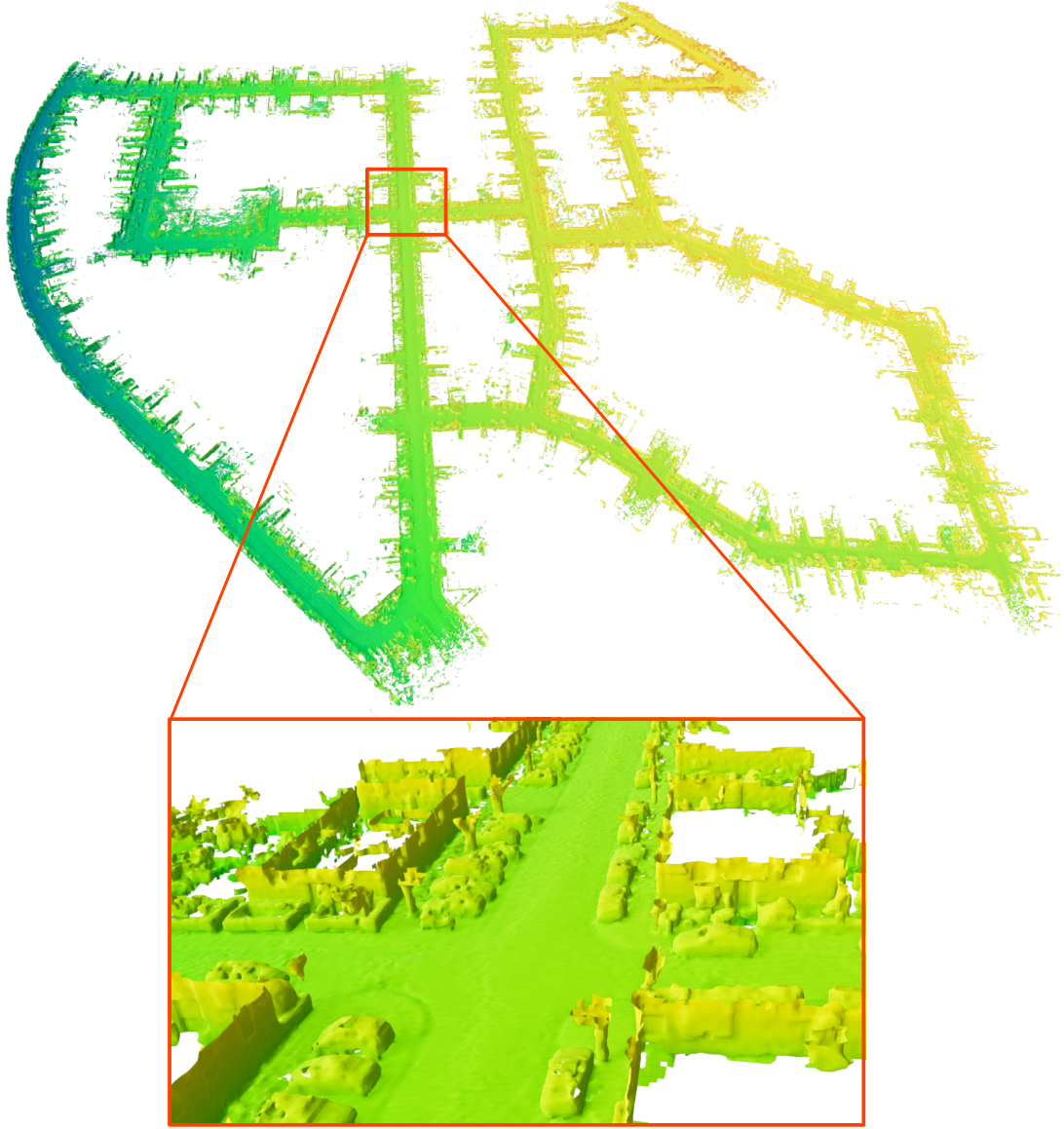}
  \caption{The incremental reconstruction result of the proposed approach on KITTI odometry sequence 00.}
  \label{fig:teaser}
 \vspace{-6pt}
\end{figure}

\begin{figure*}[ht]
	\centering
	\includegraphics[width=0.86\linewidth]{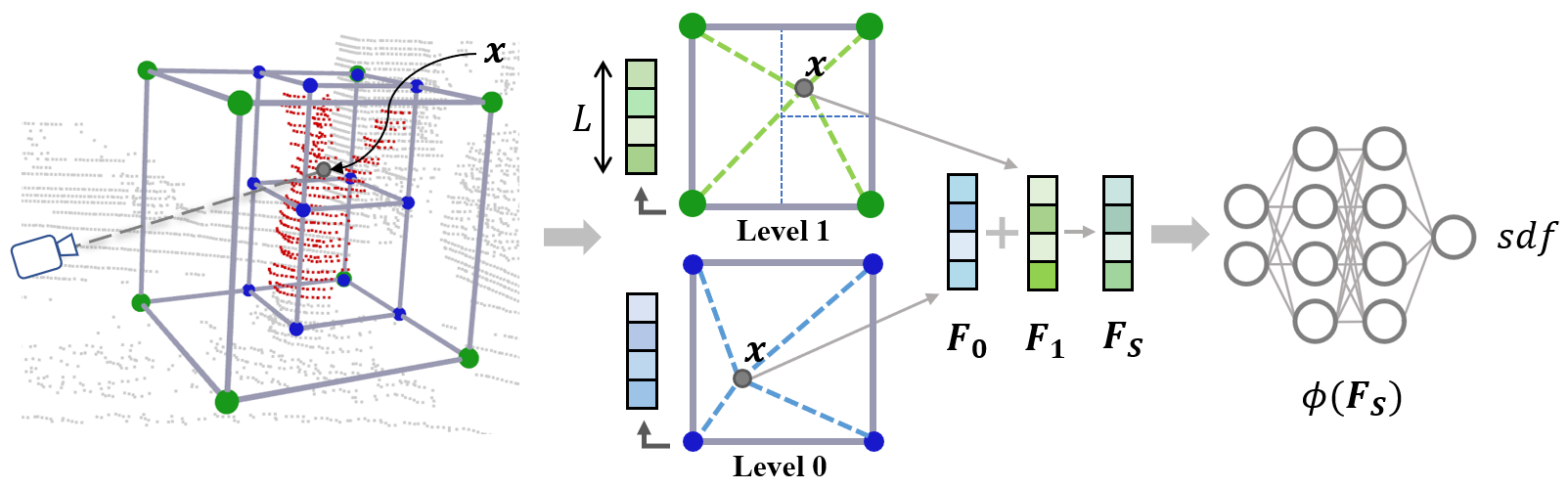}
  \setlength{\abovecaptionskip}{1pt}
	\caption{Overview of querying for SDF value in our map representation.  For querying a SDF value, we first determine the grids in each level with features where the query point $\boldsymbol{x}$ belongs to and determine the interpolated feature at the location of the query point by trilinear interpolation of the vertex features. We add the interpolated features of every level as $\boldsymbol{F_s}$ to regress the SDF value using the MLP $\phi$.}
	\label{fig:pipeline}
	\vspace{-12pt}
\end{figure*}
% TODO the notations are not unified throughout the methodlogy section (Change)
% TO change

The main contribution of this paper is a novel approach called SHINE-Mapping that enables large-scale incremental 3D mapping and allows for accurate and complete reconstructions exploiting sparse hierarchical implicit representation. An example illustration of an incremental mapping result generated by our approach on the KITTI dataset~\cite{geiger2012cvpr} is given in \figref{fig:teaser}. Our approach exploits an octree-based sparse data structure to store incrementally learnable feature vectors and a shared shallow multilayer perceptron~(MLP) as the decoder to transfer local features to signed distance values. We design a binary cross entropy-based loss function for efficient and robust local feature optimization. By interpolating among local features, our representation enables us to query geometric information at any resolution. We use point clouds as observation to optimize local features online and achieve incremental mapping through the extension of the octree structure. Our approach yields an incremental mapping system based on our implicit representation and uses the feature update regularization to tackle the issue of catastrophic forgetting~\cite{mccloskey1989plm}. As the experiments illustrate, our mapping approach (i)~is more accurate in the regions with dense observations than TSDF fusion-based methods~\cite{oleynikova2017iros,vizzo2022sensors} and volume rendering-based implicit neural mapping method~\cite{zhu2022cvpr}; (ii)~is more complete in the region with sparse observations than the non-learning-based mapping methods; (iii)~generates more compact map than TSDF fusion-based methods.

The open-source implementation is available at: ~\url{https://github.com/PRBonn/SHINE_mapping}.
%Release the code: put the Github link

%%%%%%%%%%%%%%%%%%%%%%%%%%%%%%%%%%%%%%%%%%%%%%%%%%%%%%%%%%%%%%%%%%%%%%%%%%%%%%%%
\section{Related Work}
\label{sec:related}

% Discuss the main related work and cite around 15-25 papers in sum. 
% The related work section should be approx. 1 column long, assuming 
% a 6-page paper.  Structure the section in paragraphs, grouping the 
% papers, and describing the key approaches with 1-2 sentences. If 
% applicable, describe the key difference to your approach at the end 
% of each paragraph briefly. Avoid adding subsections, al least for a 
% conference paper.

% The approach by Stachniss \etalcite{stachniss2005aaai} aims at predicting \dots

The usage of LiDAR data for large-scale environment mapping is often a key component to enable localization and navigation but also enables applications involving visualization, BIMs, or augmented reality.
%
%Besides camera data, the usage of 3D LiDAR point clouds is prevalent in robotics and particularly in context of autonomous driving~\cite{behley2018rss, vizzo2022sensors, deschaud2018icra}.
%
Besides surfel-based representations~\cite{behley2018rss}, triangle meshes~\cite{chen2021icra, marton2009icra, ruetz2019icra,vizzo2021icra}, and histograms~\cite{stachniss2003iros}, octree-based occupancy representations~\cite{hornung2013ar} are a popular choice for representing the environment. 
The seminal work of Newcombe \etal~\cite{newcombe2011ismar} that enabled real-time reconstruction using truncated signed distance function (TSDF)~\cite{curless1996siggraph} popularized the use of volumetric integration methods~\cite{kuehner2020icra,oleynikova2017iros,palazzolo2019iros,vizzo2022sensors,whelan2014ijrr}. 
The abovementioned representations result in an explicit geometrical representation of the environment that can be used for localization~\cite{chen2021icra} and navigation~\cite{oleynikova2018ral}.

While such explicit geometric representations enable detailed reconstructions in large-scale environments~\cite{niessner2013siggraph, vizzo2022sensors, whelan2014ijrr}, the recent emergence of neural representations, like NeRF~\cite{mildenhall2020eccv} for novel view synthesis inspired researchers to leverage their capabilities for mapping and reconstruction~\cite{azinovic2022cvpr, huang2021cvpr, mueller2022acmgraphics, peng2020eecv, sucar2021iccv, sun2022siggraph, sun2021cvpr, takikawa2021cvpr, zhu2022cvpr}.
Such implicit representations represent the environment via MLPs that estimate a density that can be used to volumetrically render novel views~\cite{mildenhall2020eccv} or scene geometry~\cite{mueller2022acmgraphics, sucar2021iccv, zhu2022cvpr}.

For the implicit neural mapping with sequential data, iMap~\cite{sucar2021iccv}, iSDF~\cite{ortiz2022rss} and Neural RGBD~\cite{azinovic2022cvpr} follow NeRF to use a single MLP to represent the entire indoor scene. Extending these methods to large-scale mapping is impractical due to the limited model capacity. 
By combining a shallow MLP with optimizable local feature grids, NICE-SLAM~\cite{zhu2022cvpr} and Go-SURF~\cite{wang2022threedv} can achieve more accurate and faster surface reconstruction in larger scenes such as multiple rooms. 
However, the most notable shortcoming of these methods is the enormous memory cost of dense voxel structures. Our method leverages the octree-based sparse feature grid to reduce memory consumption significantly.

Additionally, incremental mapping with implicit representation can be regarded as a continual learning problem, which faces the challenge of the so-called catastrophic forgetting~\cite{mccloskey1989plm}. 
To solve this problem, iMap~\cite{sucar2021iccv} and iSDF~\cite{ortiz2022rss} replay keyframes from historical data and train the network together with current observations. 
Nevertheless, for large-scale incremental 3D mapping, such replay-based methods will inevitably store more and more data and need complex resampling mechanisms to keep it memory-efficient. 
By storing features in local grids and using a fixed MLP decoder, NICE-SLAM~\cite{zhu2022cvpr} claims it does not suffer too much from the forgetting issue.
But as the feature grid size increases, the impact of the forgetting problem becomes more severe, which would be illustrate in detail in \secref{sec:main}.
In our approach, we achieve incremental mapping with limited memory by leveraging feature update regularization.

%\vspace{-2pt}

%% BRIEFLY SUMMARIZE OWN CONTRIBUTION 

%%%%%%%%%%%%%%%%%%%%%%%%%%%%%%%%%%%%%%%%%%%%%%%%%%%%%%%%%%%%%%%%%%%%%%%%%%%%%%%%
%%%%%%%%%%%%%%%%%%%%%%%%%%%%%%%%%%%%%%%%%%%%%%%%%%%%%%%%%%%%%%%%%%%%%%%%%%%%%%%%
\section{Our Approach -- SHINE-Mapping}
\label{sec:main}
%\vspace{-2pt}
We propose a framework for large-scale 3D mapping based on an implicit neural representation taking point clouds from a range sensor such as LiDAR with known poses as input. Our implicit map representation uses learnable octree-based hierarchical feature grids and a globally shared MLP decoder to represent the continuous signed distance field~(SDF) of the environment. As illustrated in \figref{fig:pipeline}, we optimize local feature vectors online to capture the local geometry by using direct measurements to supervise the output signed distance value from the MLP. Using this learned implicit map representation, we can generate an explicit geometric representation in the form of a triangle mesh by marching cubes~\cite{lorensen1987siggraph} for visualization and evaluation.

\subsection{Implicit Neural Map Representation}

Our implicit map representation has to store spatially located features in the 3D world. The SDF values will then be inferred from these features through the neural network.  Our network will not only use features from one spatial resolution but combine features at $H$ different resolutions. The $H$ hierarchical levels would always double the spatial extend in $x,y,z$-direction from one level to the next. We found that $H$ equals 3 or 4 is sufficient for good results.
% not fix 4 level here (but it's better add a table for the parameter settings in the experiment as the implementation details)
%Through experiments, we found that 4 hierarchical levels, always doubling the spatial extend in $x,y,z$-direction from one level to the next, provides the best results.
%

\textbf{Spatial Model.} We first have to specify our data structure. To enable large-scale mapping with implicit representation, one could choose an octree-based map representation similar to the one proposed in NGLOD~\cite{takikawa2021cvpr}. In this case, one would store features per octree node corner and consider several levels of the tree when combining features to feed the network. However, NGLOD is designed to model a single object, and the authors do not target the incremental extension of the mapped area. Extending the map is vital for the online robotic operation as the robot's workspace is typically not known beforehand, especially when navigating through large outdoor scenes.

Based on the NGLOD, we build the octree from the point cloud directly and take a different approach to store our features using hash tables, one table for each level, such that we maintain $H$ hash tables during mapping. For addressing the hash tables, while still being able to find features of upper levels quickly, we can exploit unique Morton codes. Morton codes, also known as Z-order or Lebesgue curve, map multidimensional data to one dimension, i.e., the spatial hash code, while preserving the locality of the data points. This setup allows us to easily extend the map without allocating memory beforehand while still being able to maintain the $H$ most-detailed levels. Thus, this provides an elegant and efficient, i.e., average case $\mathcal{O}(1)$ access to all features.

\textbf{Features.} In our representation, we store the feature, a one-dimensional vector of length $L$,  in each corner of the tree nodes. 
The MLP will take the same length vector as the input and compute the SDF values. For that operation, the feature vectors from at most $H$ levels of our representation are combined to form the network's input. For these features stored in corners, we randomly initialize their values when created and then optimize them until convergence by using the training pairs sampled along the rays.

\figref{fig:pipeline} illustrates the overall process of training our implicit map representation, and the right-hand side of the image illustrates the combination of features from different levels. For clearer visualization, we depict only two levels, green and blue. For any query location, we start from the lowest level (level~0) and compute a trilinear interpolation for the position~$\boldsymbol{x}$. This yields the feature vector at level~0. Then, we move up the hierarchy to level~1 to repeat the process. We combine features through summation for up to $H$ levels. 

Next, we feed the summed-up feature vector into a shared shallow MLP with $M$ hidden layers to obtain the SDF value. As the whole process is differentiable, we can optimize the feature vectors and the MLP's parameters jointly through backpropagation. 
To ensure our shared MLP generalizes well at various scales, we do not stack the query point coordinates under the final feature vector as done by Takikawa \etal~\cite{takikawa2021cvpr}. 

Our MLP does not need to be pre-trained if mapping operates in batch mode, i.e., all range measurements taken at known poses are available. In case we map incrementally, we, in contrast, use a pre-trained MLP and keep it fixed to minimize the effects of catastrophic forgetting.

\subsection{Training Pairs and Loss Function}

\begin{figure}
  \centering  
  \includegraphics[width=0.92\linewidth]{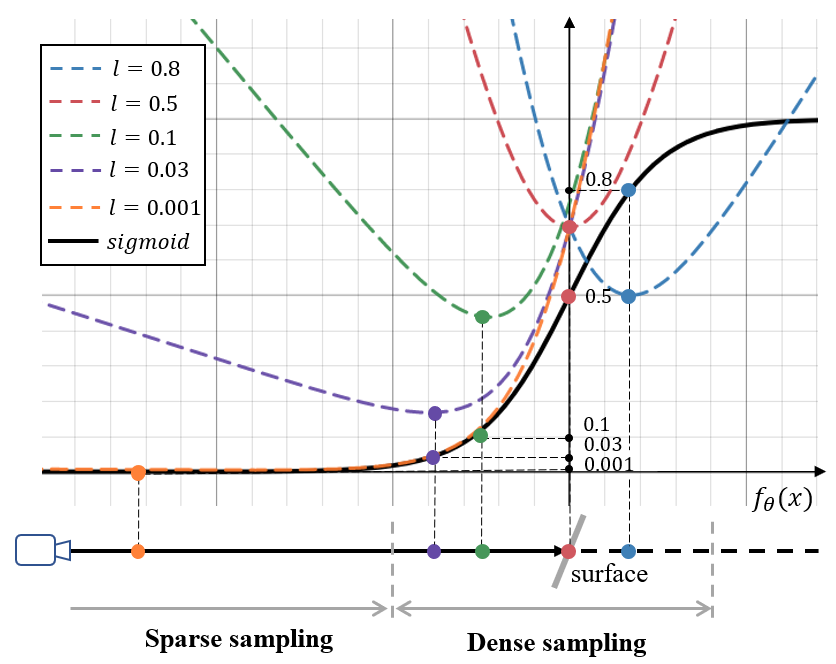}
  \setlength{\abovecaptionskip}{4pt}
  \caption{The dashed lines show the $L_\text{bce}$ curves for different labels~$l$ with network output $f_{\theta }(\boldsymbol{x})$ as horizontal coordinates. $l$ is calculated from different sample point (colorful dot) along the ray. As can be seen, for the point close to the surface, its loss is more sensitive to changes in network output. Moreover, if the network mispredicts the sign’s symbol for any sampling point, there will be a significant loss, which is what we expect. }
  \label{fig:loss_func}
  \vspace{-10pt}
\end{figure}

Range sensors such as LiDARs typically provide accurate range measurements. Thus, instead of using the volume rendering depth as the supervision as done by vision-based approaches~\cite{azinovic2022cvpr,oechsle2021iccv, wang2022threedv, zhu2022cvpr}, we can obtain more or less directly from the range readings. We obtain training pairs by sampling points along the ray and directly use the signed distance from sampled point to the beam endpoint as the supervision signal. This signal is often referred to as the projected signed distance along the ray. 

For SDF-based mapping, the regions of interest are the values close to zero as they define the surfaces. Therefore, sampled points closer to the endpoint should have a higher impact as the precise SDF value far from a surface has very little impact. Thus, instead of using an L2 loss, as, for example, used by Ortiz \etal~\cite{ortiz2022rss}, we map the projected signed distance to $[0,1]$ before feeding to the actual loss through the sigmoid function:
  $S(x) = 1/(1+e^{x/\sigma})$,
where the $\sigma$ is a hyperparameter to control the function's flatness and indicates the magnitude of the measurement noise.
% new version
Given the sampled data point $\boldsymbol{x}_{i} \in \mathbb{R}^3$, we calculate its projected signed distance to the surface $d_{i}$, then use the value after sigmoid mapping $l_{i} = S(d_{i})$ as supervision label and apply the binary cross entropy (BCE) as our loss function:
% old version
%Given the sampled data points $\left \{ \boldsymbol{x}_{1}, \boldsymbol{x}_{2},\dots,  \boldsymbol{x}_{n}  \right \}$, we calculate their projected signed distances to the surface, then use the value after sigmoid mapping $\left \{ l_{1}, l_{2},\dots,  l_{n}  \right \}$ as supervisions and apply the binary cross entropy (BCE) as our loss function:
\begin{equation}
  %\footnotesize
  \label{bce}
  L_\text{bce} = l_{i} \cdot \log(o_{i}) +  (1-l_{i}) \cdot \log(1 - o_{i}),
\end{equation}
where $o_{i} = S(f_{\theta }(\boldsymbol{x}_{i}))$ represents the signed distance output of our model $f_{\theta }(\boldsymbol{x}_{i})$ after the sigmoid mapping, which can be regarded as an indicator for the occupancy probability. The effect of BCE loss is visualized in \figref{fig:loss_func}. Additionally, the sigmoid function also realizes soft truncation of signed distance, and the truncation range can be adjusted by changing the $\sigma$ in the sigmoid function. In order to improve efficiency, we uniformly sample $N_f$ points in the free space and $N_s$ points inside the truncation band $\pm 3\sigma$ around the surface.
% For the consideration of 

Since our network output is the signed distance value, we add an Eikonal term to the loss function to encourage accurate, signed distance values like~\cite{ortiz2022rss}. Therefore, for batch-based feature optimization training, i.e., non-online mapping,  our loss function is as follows:
\begin{equation}
  %\footnotesize
  \label{batch_loss}
     L_\text{batch} =   L_\text{bce} + \lambda_{e} \underbrace{\left (  \left \| \frac{\partial f_{\theta}\left ( \boldsymbol{x}_{i}  \right ) }{\partial \boldsymbol{x}_{i}} \right \|   -1 \right )^{2}}_{\text{Eikonal loss}},   
\end{equation} 
where $\lambda_{e}$ is a hyperparameter representing the weight for the Eikonal loss, and the gradient of the input sample can be efficiently calculated through automatic differentiation.

\subsection{Incremental Mapping Without Forgetting}

\begin{figure}
  \centering  
  \includegraphics[width=0.63\linewidth]{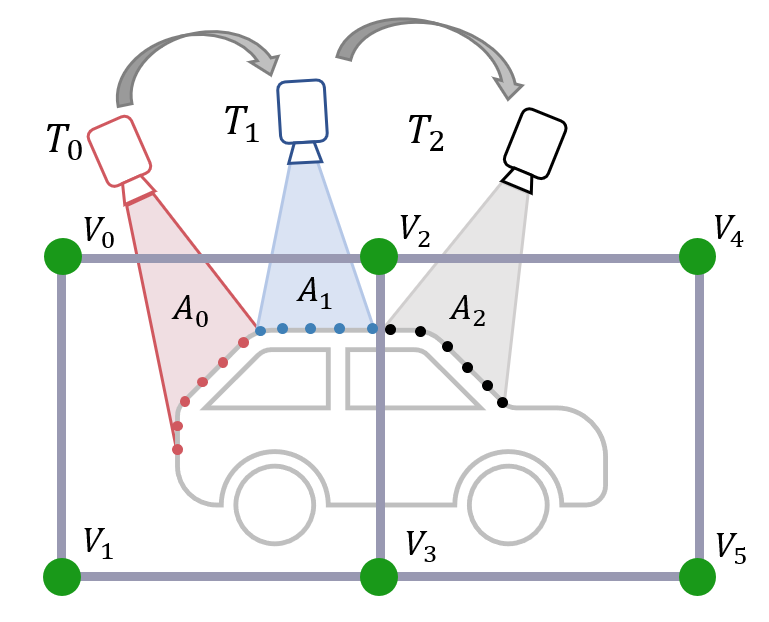}
  \setlength{\abovecaptionskip}{0pt}
  \caption{An example for the catastrophic forgetting in feature grid based incremental mapping.}
  \label{fig:forgetting_problem}
  \vspace{-10pt}
\end{figure}

As we explained in \secref{sec:related}, it is not a good choice to use the replay-based method in large-scale implicit incremental mapping. Thus, in this section, we focus on only using the current observation to train our feature field incrementally. 

Firstly, we describe the reason catastrophic forgetting happens in feature grid-based implicit incremental mapping. As shown in the \figref{fig:forgetting_problem}, we can only obtain partial observations of the environment at each frame ($T_{0}$, $T_{1}$ and $T_{2}$). 
During the incremental mapping, at frame $T_{0}$, we use the data capture from area $A_{0}$ to optimize the feature $V_{0}$, $V_{1}$, $V_{2}$, and $V_{3}$. 
After the training converge, $V_{0}$, $V_{1}$, $V_{2}$, and $V_{3}$ will have an accurate geometric representation of $A_{0}$. 
However, if we move forward and use the data from frame $T_{1}$ to train and update $V_{0}$, $V_{1}$, $V_{2}$, and $V_{3}$, the network will only focus on reducing the loss generated in $A_{1}$ and does not care about the performance in $A_{0}$ anymore.
This may lead to a decline in the reconstruction accuracy in $A_{0}$, which is why catastrophic forgetting happens. 
In addition, when we continue to use the observation at frame $T_{2}$ for training, the feature $V_{2}$, $V_{3}$ will be updated again, and we cannot guarantee that the feature update caused by the data from $T_{2}$ would not deteriorate the mapping accuracy in $A_{0}$ and $A_{1}$. \figref{fig:incrementalR} shows the impact of this forgetting problem during incremental mapping, which will become more severe as the grid size increases.

An intuitive idea to solve this problem is to limit the update direction of the local feature vector to make current update not affect the previously modeled area excessively. Inspired by the regularization-based methods in the continual learning field \cite{aljundi2018eccv, kirkpatrick2017pans, zenke2017icml}, we add a regularization term to the loss function:
\begin{equation}
  \label{Lr}
 L_\text{r} =  \sum_{i\in A} \Omega _{i} (\theta_{i}^{t} - \theta_{i}^{*})^{2},
\end{equation}
where $A$ refers to the set of local feature parameters that are used in this iteration, $\theta_{i}^{t}$ represent the current parameter values, and $\theta_{i}^{*}$ represent parameter values, which have converged in the training of previous scans. The term $\Omega _{i}$ act as the importance weights for different parameters. 

In our case, we regard these importance weights as the sensitivity of the loss of previous data to a parameter change, suggested by Zenke \etal~\cite{zenke2017icml}.
After the convergence of each scan's training, we recalculate the importance weights by:
\begin{equation}
  %\footnotesize
  \label{importance weight}
 \Omega_{i} = \min \left ( \Omega_{i}^{*} + \sum_{k=1}^{N}\left \|\frac{\partial L_\text{bce}(\boldsymbol{x}_{k},l_{k}) }{\partial \theta_{i} }   \right \|, \Omega_{m}  \right ),
\end{equation}
where $\Omega_{i}^{*}$ represents the previous importance value, and $\Omega_{m}$ is used to limit the weight to prevent gradient explosion. The iteration from $k=1$ to $k=N$ means that all samples are considered in the training of this scan. 

Intuitively, the gradient of the loss $L_\text{bce}$ to $\theta_{i}$ indicates the change of the loss on previous data by adjusting the parameter $\theta_{i}$. 
In the following training, we prefer changing the parameters with small importance weights to avoid a significant impact on previous losses. 
Therefore, for incremental mapping, our complete loss function is:
\begin{equation}
  \label{loss}
L_\text{incr} = L_\text{bce} + \lambda_{e} L_\text{eikonal} + \lambda_{r} L_{r},
\end{equation}
where $\lambda_{r}$ is a hyperparameter used to adjust the degree of keeping historical knowledge.
As our experiments will show, this approach reduces the risk of catastrophic forgetting.

%%%%%%%%%%%%%%%%%%%%%%%%%%%%%%%%%%%%%%%%%%%%%%%%%%%%%%%%%%%%%%%%%%%%%%%%%%%%%%%%
\section{Experimental Evaluation}
\label{sec:exp}

The main focus of this work is an incremental and scalable 3D mapping system using a sparse hierarchical feature grid and a neural decoder. In this section, we analyze the capabilities of our approach and assess its performance. 
%The experiments show that our method has advantages in the regions with dense observations by getting a high accuracy.
%in the region with sparse observations as it can (partially) complete the scan; in the regions without observations as by providing reasonable guesses about the 3D structure. 

\begin{figure*} 
  \centering
  \includegraphics[width=0.95\linewidth]{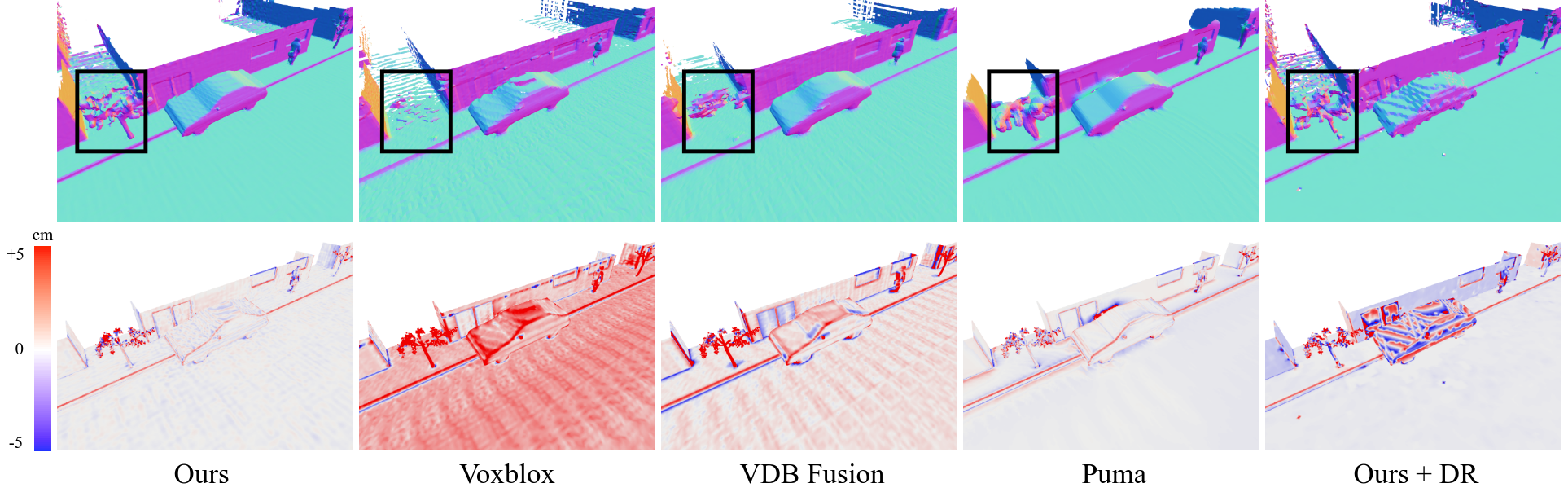}
  \setlength{\abovecaptionskip}{2pt}
  \caption{A comparison of different methods on the \emph{MaiCity dataset}.  The first row shows the reconstructed mesh and a tree is highlighted in the black box. The second row shows the error map of the reconstruction overlaid on the ground truth mesh, where the blue to red colormap indicates the signed reconstruction error from -5\,cm to +5\,cm.}
  \label{fig:experiments_on_maicity}
  \vspace{-6pt}
\end{figure*}

\begin{table}[b]
  \vspace{-4pt}
  %\footnotesize
  \centering
  \caption{Parameter settings}
  \setlength{\belowcaptionskip}{-4pt}
  \label{tab:hyperparameter}
  \setlength{\tabcolsep}{3.5mm}{
  \begin{tabular}{cccccc}
  \toprule
  $H$ & $L$ & $M$ & $N_{s}$ & $N_{f}$ & $\sigma[m]$ 
  \\ 
  \midrule
  4 & 8 & 2 & 5 & 5 & 0.05\\
  \bottomrule
  \end{tabular}}
  \vspace{-8pt}
\end{table}
% TODO: lambda not filled yet

\subsection{Experimental Setup}

Our model is first evaluated qualitatively and quantitatively on two publicly available outdoor LiDAR datasets that come with (near) ground truth mesh information.
One is the synthetic MaiCity dataset~\cite{vizzo2021icra}, which consists of a sequence of 64 beam noise-free simulated LiDAR scans of an urban scenario. The other is the more challenging, non-simulated Newer College dataset~\cite{ramezani2020iros}, recorded at Oxford University using handheld LiDAR with cm-level measurement noise and substantial motion distortion. Near ground truth data is available from a terrestrial scanner. On these two datasets, we evaluate the mapping accuracy, completeness, and memory efficiency of our approach and compare the results to those of previous methods. 
Second, we use the KITTI odometry dataset~\cite{geiger2012cvpr} to validate the scalability for incremental mapping. Finally, we showcase that our method can also achieve high-fidelity 3D reconstruction indoors. In all our experiments, we set the parameters as \tabref{tab:hyperparameter}.
 % when using scene completion.

%% Quantitative table (MaiCity)
\begin{table}[t]
  \caption{Quantitative results of the reconstruction quality on the \emph{MaiCity dataset}. We report the distance error metrics, namely completion, accuracy and Chamfer-L1 in cm. Additionally, we show the completion ratio and F-score in $\%$ with a 10\,cm error threshold.}
  \setlength{\belowcaptionskip}{-4pt}
  \centering
  \setlength{\tabcolsep}{1.2mm}{
  \begin{tabular}{lccccc}
  \toprule
  \textbf{Method}     & \textbf{Comp.~$\downarrow$}         & \textbf{Acc.~$\downarrow$}           & \textbf{C-l1~$\downarrow$}  &\textbf{Comp.Ratio~$\uparrow$}  & \textbf{F-score~$\uparrow$}     \\ \midrule
  Voxblox    & 7.1          & 1.8          & 4.8        &  84.0   & 90.9          \\
  VDB Fusion & 6.9          & 1.3          & 4.5        &  90.2   & 94.1          \\
  Puma       & 32.0         & 1.2 & 16.9        &  78.8  & 87.3          \\ \midrule
  Ours + DR & 3.3          & 1.5          & 3.7        &  94.0   & 90.7          \\
  %Go-Surf*   & 28.0         & 12.7         & 20.4         & 18.2          \\
  Ours       & \textbf{3.2} & \textbf{1.1}          & \textbf{2.9} &  \textbf{95.2}  &  \textbf{95.9} \\ \bottomrule
  \end{tabular}
  }
  \label{tab:experiments_on_maicity}
  \vspace{-4pt}
\end{table}
%%%%%%%%%%%%%%%%%%%%%%%%
\subsection{Mapping Quality}

This first experiment evaluates the mapping quality in terms of accuracy and completeness on the MaiCity and the Newer College dataset. We compare our approach against three other mapping systems: two state-of-the-art TSDF fusion-based methods, namely Voxblox~\cite{oleynikova2017iros} and VDB Fusion~\cite{vizzo2022sensors} as well as against the Possion surface-based reconstruction system Puma~\cite{vizzo2021icra}. All three methods are scalable to outdoor LiDAR mapping, and source code is available.
With our data structure, we additionally implemented differentiable rendering that is used in recent neural mapping systems~\cite{azinovic2022cvpr,wang2022threedv,zhu2022cvpr} to supervise our map representation. We denote it as Ours + DR in the experiment results.

Although our approach can directly infer the SDF at an arbitrary position,  we reconstruct the 3D mesh by marching cubes on the same fixed-size grid to have a fair comparison against the previous methods relying on marching cubes. In other words, we regard the 3D reconstruction quality of the resulting mesh as the mapping quality. For the quantitative assessment, we use the commonly used  reconstruction metrics~\cite{mescheder2019cvpr} calculated from the ground truth and predicted mesh, namely accuracy, completion, Chamfer-L1 distance, completion ratio, and F-score. 
% Check this sentence
Instead of the unfair accuracy metric used to calculate the Chamfer-L1 distance, we report the reconstruction accuracy calculated using the ground truth mesh masked by the intersection area of the 3D reconstruction from all the compared methods.

\tabref{tab:experiments_on_maicity} lists the obtained results. When fixing the voxel size or feature grid size, here to a size of 10\,cm, our method outperforms all baselines, the non-learning as well as the learnable neural rendering-based ones. The superiority of our method is also visible in the results depicted in \figref{fig:experiments_on_maicity}. 
As can be seen in the first row, our method has the most complete and smoothest reconstruction, visible, for example, when inspecting the tree or the pedestrian. The error map depicted in the second row also backs up the claims that our method can achieve better mapping accuracy in areas with dense measurements and higher completeness in areas with sparse observations compared to the state-of-the-art approaches.  

\begin{figure*}[t]
  \centering
  \includegraphics[width=0.95\linewidth]{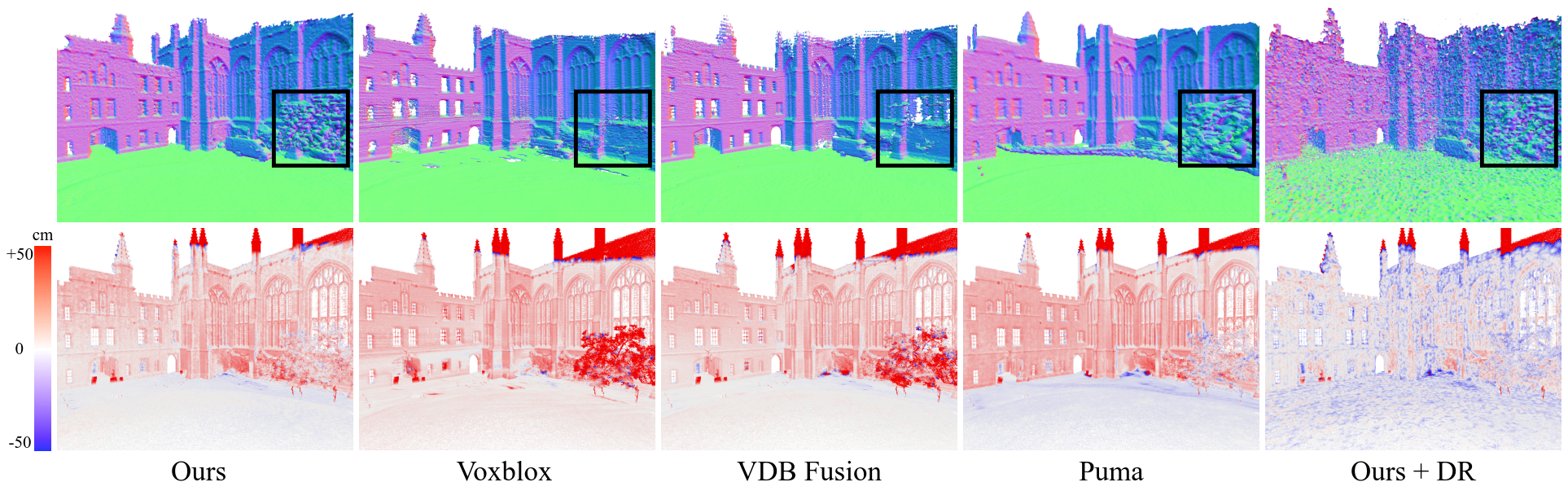}
  \setlength{\abovecaptionskip}{2pt}
  \caption{A comparison of different methods on the \emph{Newer College dataset}. The first row shows the reconstructed mesh and a tree is highlighted in the black box. The second row shows the error map of the reconstruction overlaid on the ground truth mesh, where the blue to red colormap indicates the signed reconstruction error from -50\,cm to +50\,cm.}
  \label{fig:experiments_on_ncd}
  \vspace{-14pt}
\end{figure*}

As shown in \tabref{tab:experiments_on_ncd} and \figref{fig:experiments_on_ncd}, we get better performance compared with the other methods on the more noisy Newer College dataset with the same 10\,cm voxel size. The results indicate that our method can handle measurement noise well. Besides, our approach is able to build the map, eliminating the dynamic objects such as the moving pedestrians, while not deleting the sparse static objects such as the trees. This is difficult to achieve for the space carving in Voxblox and VDB Fusion and the density thresholding in Puma.

% TBA: show the very noisy point cloud

%% Quantitative table (NCD)
\begin{table}[t]
  \caption{Quantitative results of the reconstruction quality on the \emph{Newer College dataset}. We report completion, accuracy and Chamfer-L1 in cm. Additionally, we show the completion ratio and F-score in $\%$ calculated with a 20\,cm error threshold.}
  \setlength{\belowcaptionskip}{-4pt}
  % may add two more columns for precision and completion rate
  % the first seg includes the non-learning based methods and the second seg includes the learning based methods
  % may add two rows for Voxfield and Cblox, respectively. But they are similar to Voxblox
  % add one row for our "real-time" incremental version
  \centering
  \setlength{\tabcolsep}{1.2mm}{
  \begin{tabular}{lccccc}
  \toprule
  \textbf{Method}     & \textbf{Comp.~$\downarrow$}         & \textbf{Acc.~$\downarrow$}           & \textbf{C-l1~$\downarrow$}    & \textbf{Comp.Ratio~$\uparrow$}     & \textbf{F-score~$\uparrow$}     \\ \midrule
  Voxblox    &  14.9       & 9.3        &    12.1       & 87.8  &  87.9     \\
  VDB Fusion &   12.0      &  6.9      &   9.4     & 91.3  &  92.6        \\
  Puma       &    15.4      & 7.7  &    11.5     &  89.9  & 91.9    \\ \midrule
  Ours + DR &   11.4        &  11.1      &  11.2       &   92.5&   86.1     \\
  Ours       &  \textbf{10.0}  &  \textbf{6.7}      & \textbf{8.4} & \textbf{93.6}  & \textbf{93.7} \\ \bottomrule
  \end{tabular}
  }
  \label{tab:experiments_on_ncd}
  \vspace{-4pt}
\end{table}

%%%%%%%%%%%%%%%%%%%%%%%%

\subsection{Memory Efficiency}

\begin{figure}[t]
  \centering
  \subfigure[MaiCity]{\includegraphics[width=0.23\textwidth]{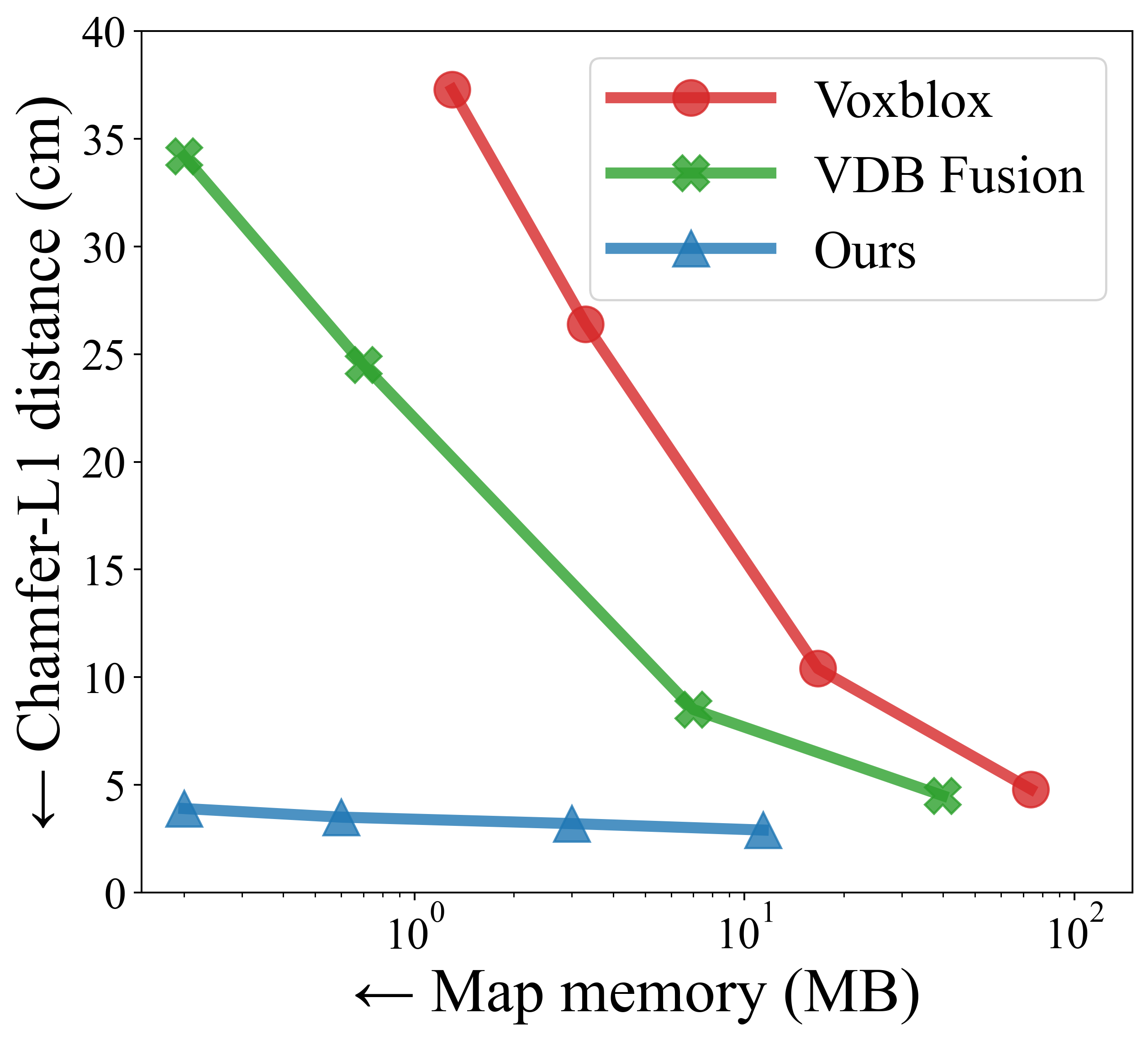}}
  \subfigure[Newer College]{\includegraphics[width=0.23\textwidth]{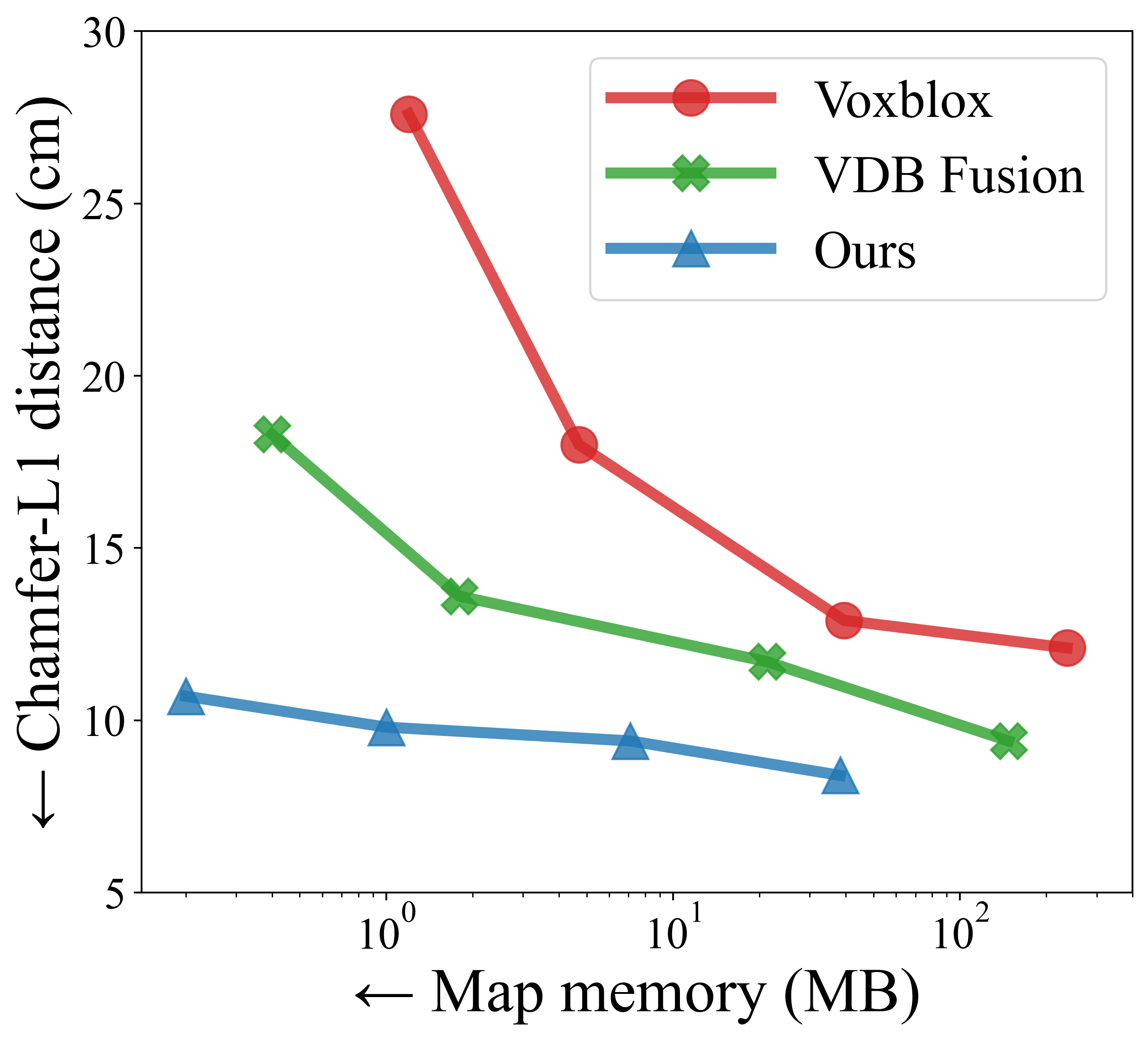}}
  \setlength{\abovecaptionskip}{0pt}
  \caption{Comparison of map memory efficiency versus the reconstruction Chamfer-L1 distance for different voxel sizes (10\,cm, 20\,cm, 50\,cm and 100\,cm). For our SHINE-Mapping, the voxel size represents the octree's leaf node grid resolution.}
  \label{fig:cd_memory}
  \vspace{-8pt}
\end{figure}

\figref{fig:cd_memory} depicts the memory usage in relation to the mapping quality. The same set of mapping voxel size settings from 10 to 100\,cm are used for the methods. The results indicate that our method can create map with smaller memory consistently for all settings. Meanwhile, our mapping quality hardly gets worse with a lower feature grid resolution, while the mapping error of Voxblox and VDB Fusion increases significantly.

Our representation using hash tables storing features allows for efficient memory usage. Although the sparse data structures are also used in Voxblox and VDB Fusion, they need to allocate memory with the same voxel size for the truncation band close to the surface and even the free space covered by the measurement rays with the space carving.

% Memory vs accuracy figure

%%%%%%%%%%%%%%%%%%%%%%%%
\subsection{Scalable Incremental Mapping}

The next experiment showcases exemplarily the ability of SHINE-Mapping to scale to larger environments, even when performing the incremental mapping. For this, we use the KITTI dataset. 
As shown in \figref{fig:teaser}, our method reconstructs a driving sequence over a distance of about 4 km with the incremental update of the hierarchical feature grids using the regularization-based continual learning strategy.  

\begin{figure}[t]
  \centering
  \subfigure[Without regularization]{\includegraphics[width=0.2\textwidth]{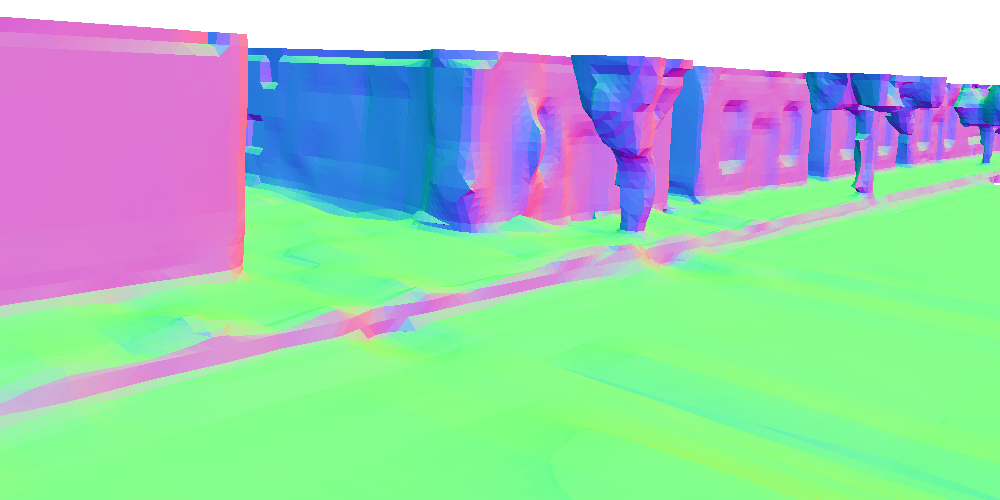}}
  \subfigure[With regularization]{\includegraphics[width=0.2\textwidth]{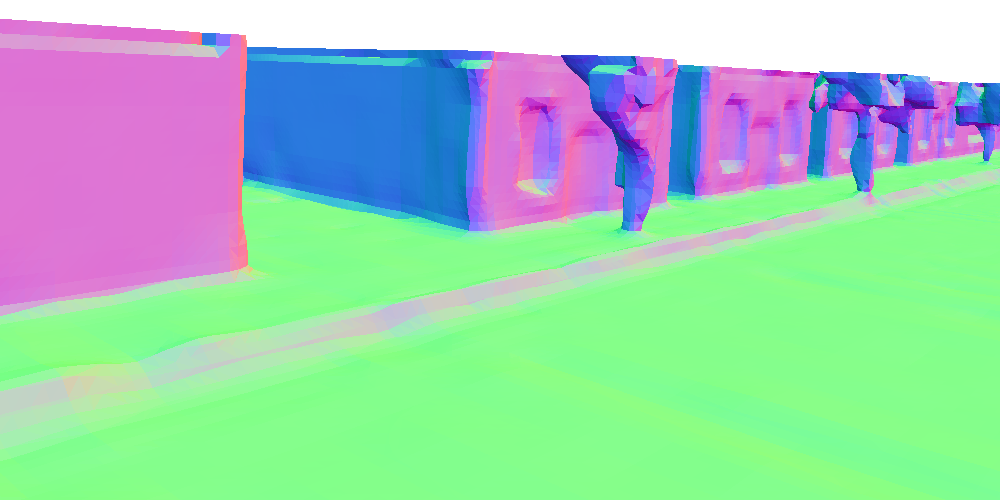}}
  \setlength{\abovecaptionskip}{0pt}
  \caption{A comparison of the incremental mapping results with and without regularization. The resolution of feature leaf nodes is 50\,cm.}
  \label{fig:incrementalR}
  \vspace{-8pt}
\end{figure}

Additionally, we provide a qualitative comparison between our incremental mapping with and without the feature update regularization in \figref{fig:incrementalR}.
The regularized approach manages to clearly reconstruct the structures that may vanish or distort as the consequences of forgetting during the incremental mapping without regularization.

%% Incremental mapping compare

% \begin{figure}[t]
% 	\centering  
% 	\includegraphics[width=0.9\linewidth]{pics/incremental_example.png}
% 	\caption{A comparison of the incremental mapping results with and without regularization. The difference are highlighted in the black boxes. The resolution of feature leaf nodes is 50\,cm.}
% 	\label{fig:incrementalR}
% 	\vspace{-10pt} % change to save space
% \end{figure}

%%%%%%%%%%%%%%%%%%%%%%%%
\subsection{Indoor Mapping and Filling Occluded Areas}

\begin{figure}[t]
  \centering
  \includegraphics[width=0.99\linewidth]{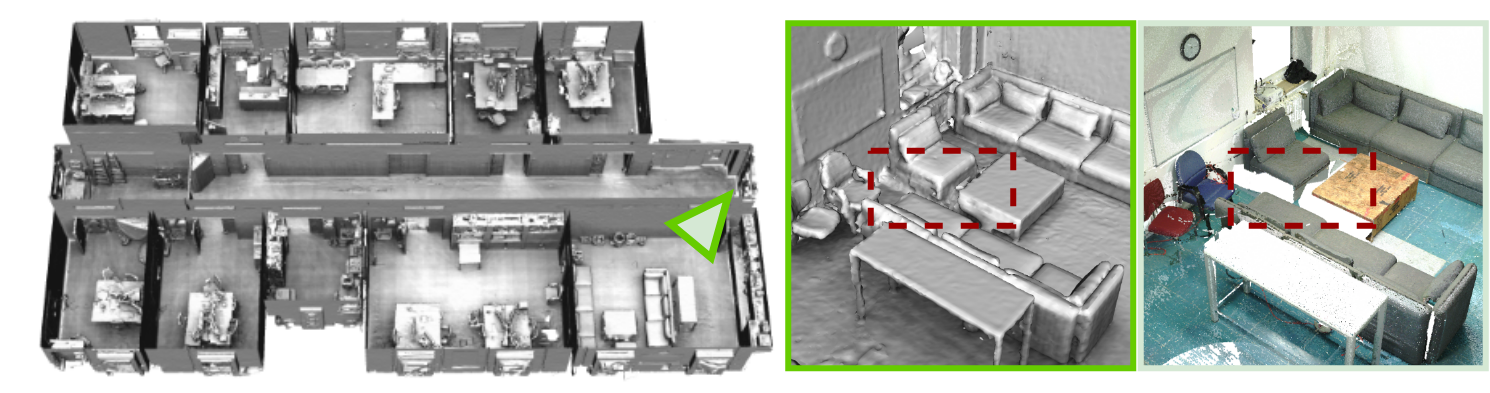}
  \setlength{\abovecaptionskip}{-8pt}
  \caption{Indoor mapping of the IPB office dataset. SHINE-Mapping's reconstruction of the whole floor is shown on the left. A close-up view of the reconstructed mesh and the input point cloud of one room are shown on the right. Our approach manages to conduct reasonable scene completion in the highlighted box.}
  \label{fig:ipb_indoor_mapping}
  \vspace{-8pt}
\end{figure}

Lastly, we provide an example for indoor mapping by our approach. \figref{fig:ipb_indoor_mapping} shows our lab in Bonn, reconstructed from sequential point clouds. We used a 3\,cm leaf node resolution instead of 10\,cm used outdoors to cover the details of indoor envrionments better. The map memory here is only 84\,MB. Our method can achieve not only the detailed reconstruction of the furnitures or the approx.~1\,cm thick whiteboard on the wall, but also a reasonable scene completion in the occluded areas where no observation are available.

\section{Conclusion}
\label{sec:conclusion}

This paper presents a novel approach to large-scale 3D SDF mapping using range sensors. Our model does not explicitly store signed distance values in a voxel grid. Instead, it uses an octree-based implicit representation consisting of features stored in hash tables and which can, through a neural network, be turned into SDF values. The network and the features can be learned end-to-end from range data. We evaluated our approach on both simulated and real-world datasets and show our reconstruction approach has advantages over current state-of-the-art mapping systems. The experiments suggest that our method achieves more accurate and complete 3D reconstruction with lower map memory than the compared methods while operating online incrementally. Furthermore, our approach can provide a reasonable guess about the structure for regions not covered by the sensor, for example, due to occlusions. 
% SDF- and Poisson-based 
% Our method provides more accurate and more complete models. 

\clearpage
\bibliographystyle{plain_abbrv}

% All new citations should go to new.bib. The file glorified.bib should go
% be the one from the ipb server. After paper or related work has been
% written merge the entries from new.bib to glorified.bib ON THE SERVER,
% replace the glorified.bib in this repository and empty the new.bib
\bibliography{glorified,new}

\end{document}